\documentclass{article}

\usepackage{microtype}
\usepackage{graphicx}
\usepackage{subcaption}
\usepackage{booktabs} 
\usepackage{pifont}   
\usepackage{hyperref}
\usepackage{enumitem} 
\usepackage{pifont}    

\usepackage{tcolorbox}
\usepackage{listings}
\usepackage{xcolor}

\usepackage[preprint]{icml2026}

\usepackage{amsmath}
\usepackage{amssymb}
\usepackage{mathtools}
\usepackage{amsthm}

\usepackage[capitalize,noabbrev]{cleveref}

\theoremstyle{plain}
\newtheorem{theorem}{Theorem}[section]

\theoremstyle{definition}
\newtheorem{definition}[theorem]{Definition}

\theoremstyle{remark}

\usepackage[textsize=tiny]{todonotes}

\icmltitlerunning{Agents on a Tree: Pathwise Coordination for Multi-Objective Molecular Optimization}

\begin{document}

\twocolumn[
  \icmltitle{Agents on a Tree: Pathwise Coordination for Multi-Objective Molecular Optimization}



  \icmlsetsymbol{equal}{*}

  \begin{icmlauthorlist}
    \icmlauthor{Jia Zhang}{yyy}
    \icmlauthor{Tengfei Ma}{yyy}
    \icmlauthor{Tianle Li}{comp}
    \icmlauthor{Daojian Zeng}{comp}
    \icmlauthor{Xieping Gao}{comp}
    \icmlauthor{Xiangxiang Zeng}{yyy}
  \end{icmlauthorlist}

  \icmlaffiliation{comp}{College of Information and Science, Hunan Normal University}
  \icmlaffiliation{yyy}{College of Computer Science and Electronic Engineering, Hunan University}
  \vspace{-0.2cm}   

  \icmlcorrespondingauthor{Daojian Zeng}{zengdj916@163.com}
  \icmlcorrespondingauthor{Tengfei Ma}{tfma@hnu.edu.cn}
    \icmlcorrespondingauthor{Xiangxiang Zeng}{xzeng@hnu.edu.cn}

  \icmlkeywords{Machine Learning, ICML}

  \vskip 0.3in
]



\printAffiliationsAndNotice{}  

\begin{abstract}

Multi-objective molecular optimization requires searching vast chemical spaces under conflicting objectives, where early design decisions strongly constrain downstream outcomes. Existing methods typically rely on a single policy or fixed scalarization, which limits their ability to represent diverse trade-offs and to explore multiple promising design trajectories.
We propose ATOM, a multi-agent framework that formulates molecular optimization as a tree-structured search. Each node corresponds to an atomic operation and hosts an agent specialized for a particular objective or decision context. Agents coordinate along different paths of the tree rather than enforcing a global consensus, enabling the method to maintain and compare alternative molecular evolution trajectories. A global memory of past optimization behaviors further supports balanced exploration and exploitation across objectives.
This tree-structured interaction enables reasoning over long-horizon dependencies inherent in molecular design. Experiments on challenging multi-objective benchmarks involving activity, synthesizability, and ADMET-related properties show that ATOM consistently achieves improved Pareto coverage and hypervolume over strong baselines. These results demonstrate the effectiveness of pathwise multi-agent coordination for molecular optimization. Code is available at https://anonymous.4open.science/r/ATOM-41CE.
\end{abstract}

\section{Introduction}
\begin{figure}[htbp]
  \centering
  \includegraphics[width=1\linewidth]{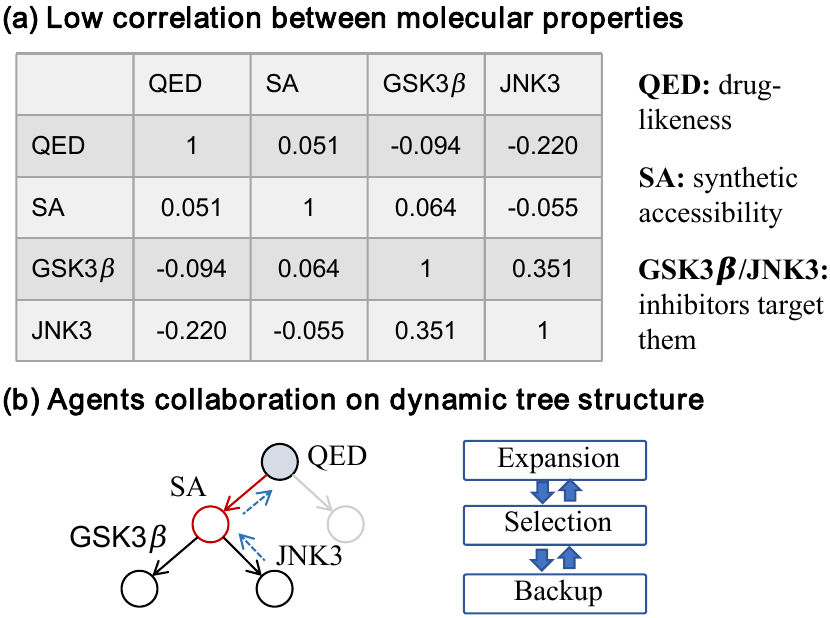}
  \caption{(a) Correlations between representative molecular properties are weak or conflicting, illustrating the intrinsic difficulty of balancing multiple objectives. (b) Our tree-structured framework coordinates specialized agents along different search paths, enabling the exploration of alternative molecular evolution trajectories without enforcing a single global policy.}
  \label{fig:example}
\end{figure}

Multi-objective molecular optimization is a central task in early-stage drug discovery~\cite{de2018challenges,yang2024enabling,liu2025multi}, where the goal is to refine lead compounds to simultaneously satisfy multiple, often conflicting, properties such as bioactivity, drug-likeness, and synthesizability~\cite{he2021molecular}. Traditional approaches, including high-throughput screening (HTS)~\cite{graff2021accelerating} and simulation-based methods~\cite{hsu2017integrated}, are effective but typically require substantial time and computational resources, limiting their scalability.

Driven by advances in artificial intelligence, machine learning has emerged as a powerful paradigm for accelerating molecular discovery~\cite{hoffman2022optimizing}. Many existing methods for multi-objective molecular design reduce the problem to a single-objective formulation by assigning predefined weights to individual objectives~\cite{maziarka2020mol,ji2021graph,xia2024evolutionary}. While this strategy can be effective in practice, it relies heavily on expert-designed weightings that are often difficult to calibrate and may bias the search toward suboptimal trade-offs when objectives are strongly conflicting~\cite{xie2021mars,fromer2023computer}. Alternatively, Pareto-based approaches attempt to approximate the Pareto front through large-scale sampling followed by non-dominated sorting~\cite{yasonik2020multiobjective,verhellen2022graph}. However, such two-stage pipelines are computationally expensive and scale poorly as the number of objectives and candidate molecules increases.
To improve sample efficiency, Bayesian optimization and Monte Carlo Tree Search (MCTS)-based methods have been widely adopted for de novo molecular generation and multi-objective property optimization~\cite{yang2024enabling,southiratncombimots,xie2021mars,gao2022sample}. Despite their principled treatment of uncertainty, these methods often suffer from scalability issues in high-dimensional chemical spaces, as well as the computational overhead associated with Gaussian process inference or deep tree expansions. These limitations hinder their practical deployment in realistic multi-objective molecular design settings.

More recently, the rapid progress of large language models (LLMs)~\cite{openai2024gpt,bai2023qwen,dubey2024llama} has sparked growing interest in their application to molecular generation~\cite{brahmavar2024generating,wang2025bridging} and optimization~\cite{yu2025collaborative,ye2025drugassist}. LLMs provide a flexible and scalable framework for goal-conditioned generation and reasoning across heterogeneous molecular properties~\cite{nguyen2024lico}. However, existing LLM-based approaches typically formulate multi-objective optimization as a monolithic generation problem, lacking explicit mechanisms for coordinating trade-offs among conflicting objectives~\cite{liu2025amopo}. 

In this work, we view multi-objective molecular optimization not as learning a single optimal policy, but as coordinating multiple specialized decision-makers along distinct optimization paths. Based on this perspective, we propose ATOM (\textbf{A}gents on a \textbf{T}ree for multi-\textbf{O}bjective \textbf{M}olecular optimization), a multi-agent framework that formulates molecular optimization as a tree-structured search. Each node corresponds to an atomic-level operation on a molecular population and hosts an agent specialized for a particular objective or decision context. Agents coordinate pathwise along different branches of the tree rather than enforcing global consensus, enabling explicit comparison of alternative molecular evolution trajectories.
To support long-horizon coordination, ATOM incorporates a global memory that aggregates historical optimization behaviors and high-quality candidates across paths. This shared context preserves agent specialization while balancing exploration and exploitation under competing objectives. The resulting tree-structured interaction facilitates long-horizon reasoning and credit assignment in path-dependent chemical spaces.


In summary, our contributions are as follows:
(i) We introduce a pathwise, tree-structured formulation of multi-objective molecular optimization that explicitly models alternative molecular evolution trajectories.
(ii) We propose ATOM, a multi-agent framework in which specialized agents coordinate along different paths rather than collapsing into a single global policy.
(iii) We demonstrate empirically and theoretically that this structure leads to superior Pareto coverage and hypervolume on challenging multi-objective benchmarks.

\section{2. Related Work}

\subsection{Molecular Optimization}
Molecular optimization is a core problem in drug discovery and materials science, and has gradually shifted from manual experimentation to data-driven computational methods \cite{gao2022sample}. Existing approaches can be broadly categorized into two classes.
(1) Combinatorial Optimization. Traditional approaches treat molecular design as a search problem over a discrete, exponentially large chemical space \cite{bohacek1996art,stumpfe2012exploring}. Common techniques include Monte Carlo Tree Search (MCTS) \cite{yang2023large}, Genetic Algorithms (GA) \cite{jensen2019graph,fu2022reinforced}, and Reinforcement Learning (RL) \cite{bohacek1996art,stumpfe2012exploring}. While these methods explore the structural space iteratively, they often struggle with high-dimensional search landscapes and the prohibitive computational cost of evaluating complex objectives.
(2) Generative Models in Molecular Design. To mitigate the challenges of discrete search, recent research has moved toward generative modeling \cite{du2024machine}. These models learn the implicit probability distribution of chemical data to propose valid molecular candidates, effectively concentrating the search space. Various architectures have been explored, including Variational Autoencoders (VAEs) \cite{gomez2018automatic,jin2018junction}, Generative Adversarial Networks (GANs) \cite{guimaraes2017objective}, flow-based models \cite{shi2020graphaf}, and diffusion models \cite{hoogeboom2022equivariant,schneuing2024structure}.

\subsection{LLMs for Molecular Optimization}
Large language models have recently been applied to molecule-centered tasks, including property prediction and generation \cite{luo2022biogpt,li2023chatdoctor,han2023medalpaca,fang2023mol,wu2024pmc} . Several works adapt LLMs for optimization: MOLLEO uses LLMs as genetic operators to improve crossover and mutation \cite{wang2024efficient}; LICO applies context-aware prompting for in-context molecule refinement without retraining \cite{nguyen2024lico}; DrugAssist presents a human-in-the-loop optimization framework that combines human insight with LLM reasoning \cite{ye2025drugassist}.
Despite these advances, existing LLM-based approaches typically treat multi-objective optimization as a single unified generation task or instantiate agents for individual properties without mechanisms to coordinate across objectives. Consequently, inter-objective conflicts are seldom modeled explicitly and these methods cannot reliably discover coordinated, globally effective optimization trajectories. In contrast, our work frames each objective as an autonomous chemistry-aware agent and uses Monte–Carlo Tree Search to plan coordinated agent actions, enabling explicit trade-off reasoning and automated discovery of optimization paths in molecular design.

\section{Preliminary}

We study multi-objective molecular optimization over a discrete chemical space.
Let $\mathcal{X}$ denote the set of all chemically valid molecules, where each molecule
$x \in \mathcal{X}$ is represented by a valid SMILES string.
Each molecule $x$ is associated with a $K$-dimensional objective vector
$\mathbf{f}(x) = [f_1(x), \ldots, f_K(x)] \in \mathbb{R}^K$, where each objective function
$f_k : \mathcal{X} \rightarrow \mathbb{R}$ evaluates a molecular property of interest,
such as target-specific bioactivity, drug-likeness (QED), or synthetic accessibility (SA).
These scoring functions are generally non-differentiable and treated as black-box
evaluators.
Given an initial molecule $x_0 \in \mathcal{X}$, the goal is to generate a set of candidate
molecules that jointly optimize the objectives. This is formulated as a $K$-objective
maximization problem:
\begin{equation}
\max_{x \in \mathcal{X}} \; \mathbf{f}(x).
\label{eq:multi_objective}
\end{equation}
Due to conflicting objectives, no single solution is optimal for all criteria.
Instead, we aim to identify a diverse set of trade-off solutions.

\begin{definition}[Pareto Dominance]
\label{def:moo}
Let $\mathcal{S} \subset \mathbb{R}^K$ denote a non-empty set of objective vectors obtained
from candidate molecules. For any $X, Y \in \mathcal{S}$, we say that $Y$ \emph{dominates}
$X$, denoted by $Y \succ X$, if
\begin{equation}
Y \succ X
\quad \Longleftrightarrow \quad
\left\{
\begin{aligned}
& Y_k \ge X_k, \quad \forall k \in \{1, \ldots, K\}, \\
& \exists k' \in \{1, \ldots, K\} \ \text{s.t.} \ Y_{k'} > X_{k'} .
\end{aligned}
\right.
\label{eq:dominance}
\end{equation}
An objective vector $X \in \mathcal{S}$ is \emph{non-dominated} if there exists no
$Y \in \mathcal{S}$ such that $Y \succ X$.
\end{definition}

\begin{definition}[Pareto Fronts]
\label{def:pf}
The \emph{first Pareto front}, also known as the \emph{Pareto optimal set}, consists of all
non-dominated solutions:
\begin{equation}
\mathcal{S}_1
=
\left\{
X \in \mathcal{S} \;:\;
\nexists Y \in \mathcal{S} \ \text{s.t.} \ Y \succ X
\right\}.
\label{eq:pareto_front_1}
\end{equation}
Subsequent Pareto fronts are defined recursively by removing solutions in the preceding
fronts. The $k$-th Pareto front is given by
\begin{equation}
\mathcal{S}_k
=
\left\{
X \in \mathcal{S} \setminus \bigcup_{i=1}^{k-1} \mathcal{S}_i
\;:\;
\nexists Y \in \mathcal{S} \setminus \bigcup_{i=1}^{k-1} \mathcal{S}_i
\ \text{s.t.} \ Y \succ X
\right\}.
\label{eq:pareto_front_k}
\end{equation}
\end{definition}

\section{Method}
\begin{figure*}[t]
  \centering
  \includegraphics[width=0.9\textwidth]{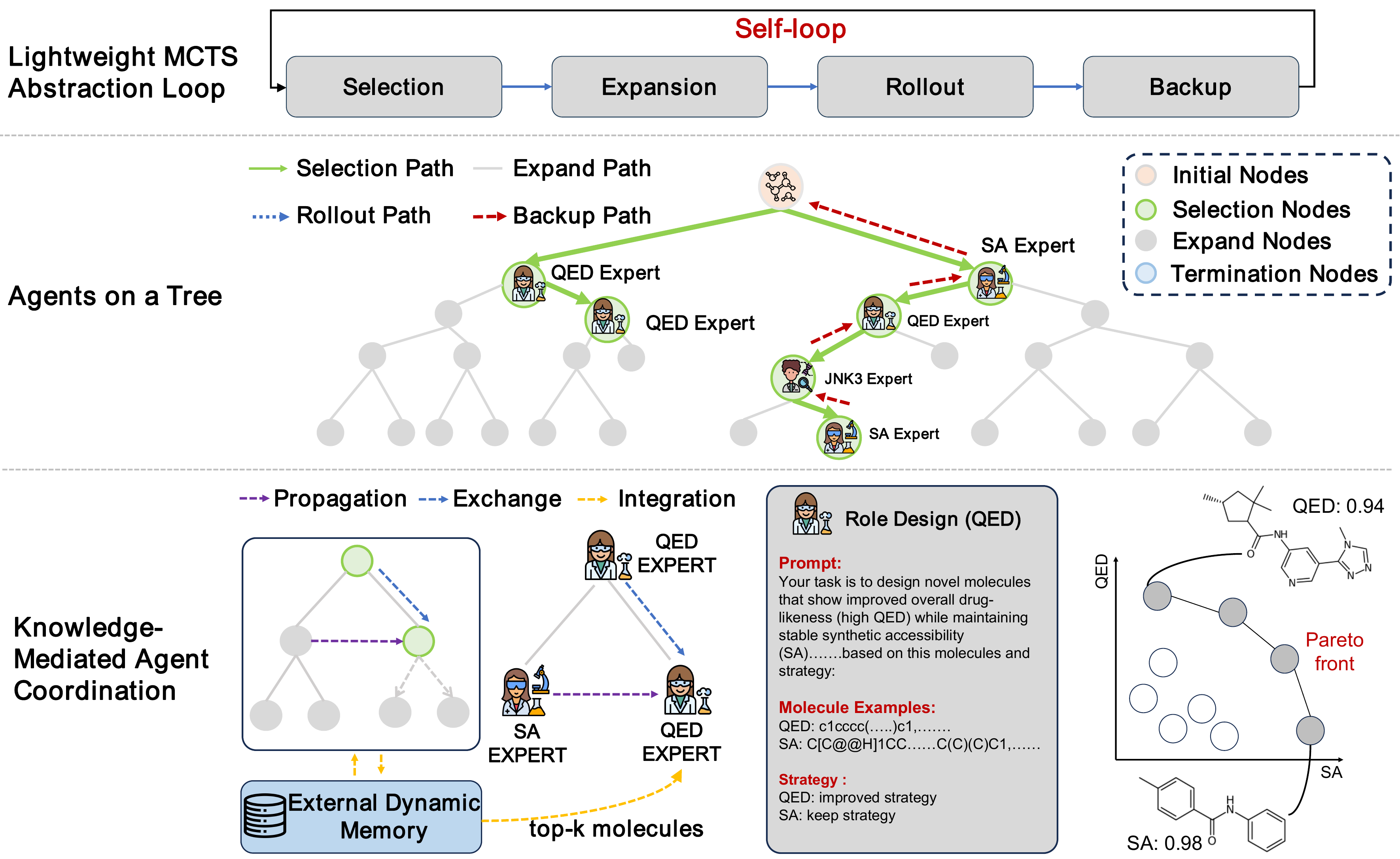}
  \caption{Agents-on-a-Tree framework for multi-objective molecular optimization, combining pathwise MCTS planning with knowledge-mediated coordination among specialized agents to improve Pareto coverage under conflicting objectives.}
  \label{fig2:framework}
\end{figure*}

\subsection{Algorithmic Framework of ATOM}
\noindent\textbf{Multi-Agent Attribute-Specific Optimization.}
As illustrated in Figure 1, we instantiate a collection of expert agents within the proposed ATOM framework, where each agent is explicitly specialized for optimizing a particular molecular property. 
Each expert is instantiated as a LLM, such as GPT-4o mini \cite{openai_gpt4omini_2024}, and assigned a well-defined domain role corresponding to a specific optimization objective. 
Concretely, these roles include a QED expert, a SA expert, and target-specific experts for GSK3$\beta$ and JNK3, which are closely associated with Alzheimer’s disease.

Recent studies have shown that LLM performance on biochemical and molecular reasoning tasks can be substantially enhanced through domain-aware prompt engineering \cite{luo2025leveraging,li2025large}. 
Motivated by these findings, ATOM adopts expert-specific prompt templates that explicitly specify the task scope, optimization objective, input representation, and expected output format for each agent. 
This design enforces functional disentanglement across experts while enabling focused and interpretable decision-making.

To further improve specialization and practical effectiveness, each expert agent in ATOM is equipped with tool-calling capabilities, enabling on-demand access to domain-specific tools such as RDKit and learned oracles for molecular property evaluation. 
These tools allow agents to ground their reasoning in quantitative feedback, thereby supporting informed optimization actions rather than purely textual heuristics. Detailed descriptions of the tool interfaces and invocation protocols are provided in Appendix A.

\noindent\textbf{Adaptive Trajectory Selection via UCT-Style Scoring.}
We employ a Monte Carlo Tree Search (MCTS) framework to adaptively select optimization trajectories in the molecular search space. Each node in the tree represents a population of molecules, and its value is designed to reflect both task-oriented optimization quality and multi-objective structural diversity. Specifically, the intrinsic value of a node $N$ is defined as
\begin{equation}
V(N) = \lambda \cdot S_{\mathrm{attr}}(N) + (1-\lambda) \cdot \widehat{\mathrm{HV}}(N),
\end{equation}
where $S_{\mathrm{attr}}(N)$ denotes an attribute-weighted score over the top-performing molecules in the node, $\widehat{\mathrm{HV}}(N)$ is the normalized hypervolume of the Pareto front induced by the node, and $\lambda \in [0,1]$ controls the trade-off between directional optimization and diversity preservation. The attribute score is computed as
\begin{equation}
S_{\mathrm{attr}}(N) =
\frac{1}{|\mathrm{Top}\text{-}k(N)|}
\sum_{m \in \mathrm{Top}\text{-}k(N)}
\sum_{i=1}^{K} w_i \tilde f_i(m),
\end{equation}
where $\tilde f_i(m)$ denotes the normalized value of the $i$-th molecular property and $w_i$ represents its corresponding importance weight provided by expert agents.

During the selection phase, child nodes are chosen by maximizing a UCT-style score:
\begin{equation}
\mathrm{UCT}(N) = V(N) + c \sqrt{\frac{\log N_{\mathrm{parent}}}{N_{\mathrm{visit}}(N)}},
\end{equation}
where $N_{\mathrm{visit}}(N)$ denotes the number of visits to node $N$, and $c$ controls the degree of exploration. This formulation prioritizes nodes with high intrinsic value while explicitly encouraging exploration of under-visited regions.

After rollout termination, node values are propagated upward using a hierarchical averaging rule:
\begin{equation}
V_{\mathrm{parent}} \leftarrow
\frac{1}{2}
\left(
V_{\mathrm{parent}} +
\frac{1}{|\mathcal{C}|}
\sum_{i=1}^{|\mathcal{C}|} V_{\mathrm{child}}^{(i)}
\right),
\end{equation}
where $\mathcal{C}$ denotes the set of child nodes. This update strategy stabilizes value estimation by integrating the parent’s prior estimate with aggregated feedback from its descendants, enabling robust and adaptive trajectory selection throughout the search process.

\subsection{Theoretical Analysis of ATOM}
The theoretical complexity of searching for optimal molecular structures typically grows exponentially with the decision depth $L$ (molecule length) and the branching factor $K$ (chemical space dimensionality), rendering exhaustive search computationally infeasible \cite{polishchuk2013estimation}. Furthermore, traditional Monte Carlo Tree Search (MCTS) is often restricted to sequential decision-making for a single molecule, which fails to capture the evolutionary characteristics of molecular populations under complex multi-objective distributions. To address these challenges, we map the molecular generation process to the ATOM framework. We propose a population-based search paradigm that leverages theoretical sample complexity bounds for value and policy networks~\cite{silver2017mastering}; the Assumptions underpinning our theoretical analysis are detailed in Appendix~F. This framework ensures that the method remains computationally tractable in high-dimensional spaces while maintaining practical relevance through population diversity.

\textbf{Escape via Agent Synergy} This step demonstrates how the synergy among specialized agents ensures the search escapes single-objective local optima by concentrating the sample budget on orthogonal descent directions. In our framework, the policy network acts as an orchestrator that assigns selection weights to $K$ specialized agents. Let the candidate child populations generated from state $s_d$ by the set of agents $\{Ag_1, \dots, Ag_K\}$ be denoted as $r_1, \dots, r_K$. According to Assumption 1, if the current population is trapped in a local optimum for objective $i$, there exists at least one orthogonal agent (denoted as $Ag_1$) that provides a significant value improvement $\delta$. Specifically, for any non-orthogonal agent $Ag_k$ ($k > 1$), we have:
\begin{equation}\mathbb{E}[V_{r_1} - V_{r_k}] \ge \delta.\end{equation}
Within the selection logic of ATOM, the expected sample complexity is governed by the algorithm's capability to identify and track this synergistic direction. The synergy manifests in the concentration of the policy distribution $p_r$: as search depth $d$ increases, the policy network adaptively assigns higher prior probabilities to specialized agents that resolve the current "property bottlenecks" of the population based on evaluation feedback. The probability that the search fails to escape the local optimum—by "mis-selecting" a sub-optimal agent $Ag_k$ that offers no marginal gain—is bounded by the likelihood that its noisy score exceeds that of the optimal agent $Ag_1$:
\begin{equation}\mathbb{P}(\text{Select } Ag_k) \approx \mathbb{P}(V_{r_1} - V_{r_k} + X^\pi_{1} - X^\pi_{k} \le \text{UCB Bias}).\end{equation}
Given the lower bound $\delta$ on the value gap from Assumption 1 and the decaying noise variance $\sigma_d$ from Assumption 2, this mis-selection probability decays at a rate of $O(\exp(-\delta^2 / \sigma_d^2))$. Consequently, the multi-agent framework ensures that the search does not stall at single-property local optima. Instead, it adaptively switches to an orthogonal agent branch, leveraging the steepest descent direction toward the Pareto front. This mechanism effectively reduces the branching factor to only those agents contributing to the joint objective improvement, thereby significantly enhancing search efficiency.

\textbf{Pruning via UCB.} \textbf{Pruning via UCB.} Expansion is restricted to nodes whose optimistic upper bounds exceed the true optimal value $V^*$, thereby pruning provably sub-optimal branches:
\begin{equation}
U(s') + c_d \ge V^* \iff V(s') + X_{s'} + c_d \ge V^*.
\label{eq7:transition} 
\end{equation}
Rearranging terms, this is equivalent to the event where the noise and exploration bonus exceed the optimality gap $\Delta_{s'} = V^* - V(s')$:
\begin{equation}
\mathbb{P}(\text{Expand } s') = \mathbb{P}(\Delta_{s'} - X_{s'} - c_d \le 0).
\label{eq6:transition} 
\end{equation}

Under Assumption 2 (Optimistic Pruning), $c_d$ is chosen such that $|X_{s'}| [cite_start]\le c_d$ with high probability.For "wrong" branches (Agent Mismatch) where the chosen agent does not address the current molecular bottleneck (e.g., optimizing QED when SA is invalid), the gap $\Delta_{s'}$ remains large.Consequently, the probability $\mathbb{P}(\Delta_{s'} \le X_{s'} + c_d)$ becomes negligible, ensuring that the number of sub-optimal expansions in the "SubOptimal" term of the main theorem remains sparse.

\textbf{Contraction via Knowledge Base} The standard sample complexity for MCTS is polynomial in the tree depth $D$ under decaying noise models.
\begin{equation}
\mathbb{E}[N] \approx O((KD)^{C}),
\label{eq6:transition} 
\end{equation}

where $C$ depends on the noise decay rate. However, in de novo molecular generation, the physical edit distance $L$ (number of atoms/bonds added) to reach a high-value molecule from scratch is large, making $K^L$ intractable.Assumption 3 asserts that conditioning on the Knowledge Base $\mathcal{K}$ allows the generation of a target molecule in $L_{KB}$ steps.This effectively replaces the physical depth $D=L$ with the effective depth $D_{eff} = L_{KB}$ in the summation of the Sample Complexity bound:

\begin{equation}
\sum_{d=0}^{L} (\dots) \longrightarrow \sum_{d=0}^{L_{KB}} (\dots).
\label{eq6:transition} 
\end{equation}
Since the complexity grows polynomially (or exponentially in worst cases) with depth, the reduction $L_{KB} < L$ provided by the KB prompts results in a significant contraction of the search space, enabling the sample complexity to remain bounded even for complex, multi-constraint objectives.

\subsubsection{Knowledge-Mediated Agent Coordination}
We introduce a knowledge-mediated coordination mechanism to enable structured information sharing among specialized expert agents within the MCTS framework. Each tree node $N$ maintains a molecular population $P_N$ and is associated with a set of expert agents $\{A_1,\dots,A_K\}$, each responsible for optimizing a specific molecular property.

\paragraph{Lateral Knowledge Exchange.}
At the same tree depth, expert agents exchange compact, attribute-ranked summaries to provide complementary guidance while preserving a designated lead objective. For node $N$, each agent $A_k$ extracts a top-ranked subset $S_k(N)$ from $P_N$ according to its objective. When agent $A_q$ is selected as the lead optimizer, its effective scoring integrates auxiliary suggestions from other agents, such as
\begin{equation}
\tilde S_q(N) = S_q(N) \cup \bigcup_{j \neq q} \omega_{j \rightarrow q} S_j(N),
\end{equation}
where $\omega_{j \rightarrow q}$ controls the influence of auxiliary agents. This mechanism enables constraint-aware optimization without diluting the primary objective.

\paragraph{Hierarchical Knowledge Propagation.}
To bias expansion toward promising regions, parent nodes propagate high-quality molecules to their children. Specifically, given a parent node $N_p$ and its Pareto front $\mathcal{P}(N_p)$, a subset of top-performing molecules is selected and injected into a child node $N_c$:
\begin{equation}
\begin{aligned}
    &\Pi(N_p \to N_c) = \text{Top-}r(\mathcal{P}(N_p)), \\
    &P_{N_c} \leftarrow P_{N_c} \cup \Pi(N_p \to N_c).
\end{aligned}
\end{equation}
This hierarchical propagation transfers favorable structural motifs and multi-objective trade-offs along the search trajectory.

\paragraph{Global Knowledge Integration.}
We further maintain a dynamic global memory $\mathcal{M}$ that aggregates high-quality molecules discovered across all search trajectories. Each node periodically contributes its best candidates to $\mathcal{M}$, while agents retrieve relevant exemplars during local optimization:
\begin{equation}
\mathcal{M} \leftarrow \mathrm{Top}\text{-}M\big(\mathcal{M} \cup \{(m, s_m)\}\big),
\end{equation}
where $s_m$ denotes a composite attribute score. Retrieved molecules are incorporated into the agent’s scoring function as global guidance:
\begin{equation}
S_{\mathrm{attr}}^{(\mathcal{M})}(N)
=
(1-\gamma)\,S_{\mathrm{attr}}(N)
+
\gamma\,S_{\mathrm{attr}}\big(\mathrm{Retrieve}(\mathcal{M})\big),
\end{equation}
with $\gamma$ controlling the influence of global knowledge.

Together, these mechanisms enable coordinated optimization across agents and search trajectories, preserving agent specialization while promoting efficient multi-objective molecular exploration.

\section{Experiments}


\textbf{Implementation Details.} We consider four molecular property objectives and analyze their pairwise correlations on a the ZINC20\cite{irwin2020zinc20} dataset.
As shown in our Figure~\ref{fig:example}, most objective pairs exhibit negative correlation, indicating substantial inter-objective conflicts.
Based on this,we design five task settings with varying conflict intensities: 1) QED+SA (non-biological objectives): Druglikeness (QED) and synthetic accessibility (SA) measure the develop ability and synthesizability of molecules, computed using RDKit. 
2) GSK3$\beta$+JNK3 (biological objectives): The inhibition of GSK3$\beta$ and JNK3, two kinase targets as sociated with Alzheimer’s disease, predicted using random forest models. 
3) QED+SA+GSK3$\beta$/JNK3: Optimization of either GSK3$\beta$ or JNK3 inhibition under constraints of good QED and SA properties.
4) GSK3$\beta$+JNK3+QED: Simultaneous optimization of GSK3$\beta$ and JNK3 inhibition with QED constraints, without explicit control over synthetic accessibility. 
5) QED+SA+GSK3$\beta$+JNK3: Joint optimization across all four objectives to balance activity, drug-likeness, and synthesizability.

\textbf{Benchmarks and Evaluation Metrics.} To ensure a diverse chemical initialization for multi-objective optimization, we leverage the ZINC20. Our optimization objective spans four critical dimensions: GSK3$\beta$ inhibition, JNK3 inhibition, QED, and SA. To rigorously quantify the performance across this multi-dimensional frontier, we adopt the hypervolume (HV) indicator \cite{zitzler2003performance} as our primary metric. HV measures the volume of the objective space dominated by the attained solution set relative to a predefined reference point, thereby providing a joint characterization of both Pareto convergence and distributional diversity. Furthermore, we supplement our analysis with standard molecular generation metrics, including novelty and diversity, to provide a holistic assessment of the generative performance.

\textbf{Baselines.} We evaluate \textbf{ATOM} against a comprehensive suite of baselines, categorized into traditional molecular optimization frameworks and recent Large Language Model (LLM)-based methods. The traditional baselines comprise: (1) \textbf{SMILES LSTM} \cite{bjerrum2017smiles}, a recurrent generative model that optimizes SMILES sequences through reinforcement learning; (2) \textbf{SMILES GA} \cite{brown2019guacamol}, which employs a genetic algorithm to evolve SMILES representations; (3) \textbf{GRAPH GA} \cite{jensen2019graph}, a genetic algorithm operating directly on molecular graph structures; (4) \textbf{STONED} \cite{nigam2021beyond}, an efficient algorithm designed for rapid chemical space exploration via random string-level mutations; and (5) \textbf{GP BO} \cite{tripp2021fresh}, a Bayesian optimization approach leveraging Gaussian processes. For LLM-based baselines, we compare against: (1) \textbf{Drugassist} \cite{ye2025drugassist}, which formulates molecule optimization as an interactive dialogue with an LLM; and (2) A direct \textbf{GPT-4o mini}\cite{achiam2023gpt}-based optimization baseline, which applies a single-agent LLM to optimize multiple objectives simultaneously (3)  \textbf{EAG} \cite{gu2025explain}, a multi-agent framework that decomposes complex optimization tasks into coordinated pipeline stages.

\subsection{Result}
\textbf{Performance Comparison on Multi-Objective Tasks.} 
Table~\ref{tab:multi_objective_results} reports the HV scores of different methods across 6 multi-objective molecular optimization tasks. 
ATOM consistently achieves the best performance in most settings, yielding the highest HV scores in each objective combination and the largest overall HV sum (4.351). 
This demonstrates its superior ability to explore diverse high-quality solutions and effectively balance conflicting objectives. 
On simpler dual-objective tasks such as QED+SA and GSK3$\beta$+JNK3, most baselines perform reasonably well, especially GRAPH GA and SMILES GA. However, their performance declines substantially as the number of objectives increases. 
Notably, SMILES LSTM and STONED perform poorly on tasks involving biological targets, highlighting their limited scalability in high-dimensional spaces. Among LLM-based approaches, single-agent methods like GPT4o-mini (HV sum: 3.678) show constrained capability in handling objective conflicts. 
While the multi-agent framework EAG (HV sum: 3.752) demonstrates improved coordination, ATOM’s specialized expert collaboration and dynamic scheduling significantly outperform all these LLM-based alternatives.

\begin{table*}[t]
\centering
\caption{Performance comparison on multi-objective molecular optimization tasks.
Higher values indicate better performance.}
\label{tab:multi_objective_results}
\setlength{\tabcolsep}{6pt}  
\renewcommand{\arraystretch}{1.1}  
{\scriptsize   
\begin{tabular}{lccccccc}
\toprule
Method
& QED+SA
& JNK3+GSK3$\beta$
& JNK3+QED+SA
& GSK3$\beta$+QED+SA
& QED+JNK3+GSK3$\beta$
& QED+JNK3+GSK3$\beta$+SA
& Sum(↑) \\
\midrule
SMILES LSTM  & 0.925±0.004 & 0.592±0.035 & 0.512±0.025 & 0.706±0.037 & 0.445±0.061 & 0.438±0.031 & 3.618 \\
SMILES GA    & 0.885±0.003 & 0.706±0.014 & 0.510±0.007 & 0.655±0.070 & 0.609±0.021 & 0.424±0.034 & 3.788 \\
GRAPH GA     & 0.899±0.002 & 0.725±0.059 & 0.582±0.072 & 0.704±0.035 & 0.610±0.071 & 0.449±0.096 & 3.969 \\
STONED        & 0.876±0.001 & 0.689±0.023 & 0.497±0.035 & 0.623±0.027 & 0.603±0.037 & 0.411±0.041 & 3.698 \\
GP BO          & 0.843±0.003 & 0.783±0.043 & 0.505±0.045 & 0.629±0.030 & 0.640±0.103 & 0.422±0.101 & 3.822 \\
Drugassist    & 0.874±0.002 & 0.645±0.011 & 0.581±0.029 & 0.692±0.047 & 0.501±0.021 & 0.484±0.061 & 3.776 \\
GPT4o-mini   & 0.880±0.001 & 0.621±0.033 & 0.583±0.025 & 0.671±0.024 & 0.499±0.013 & 0.433±0.073 & 3.678 \\
EAG           & 0.895±0.001 & 0.595±0.021 & 0.551±0.037 & 0.711±0.061 & 0.519±0.049 & 0.482±0.071 & 3.752 \\
\midrule
\textbf{ATOM} & \textbf{0.941±0.001} & \textbf{0.790±0.017} & \textbf{0.608±0.045} & \textbf{0.764±0.051} & \textbf{0.644±0.062} & \textbf{0.606±0.041} & \textbf{4.351} \\
\bottomrule
\end{tabular}
} 
\end{table*}
\begin{figure}[htbp]
  \centering
  \includegraphics[width=1\linewidth]{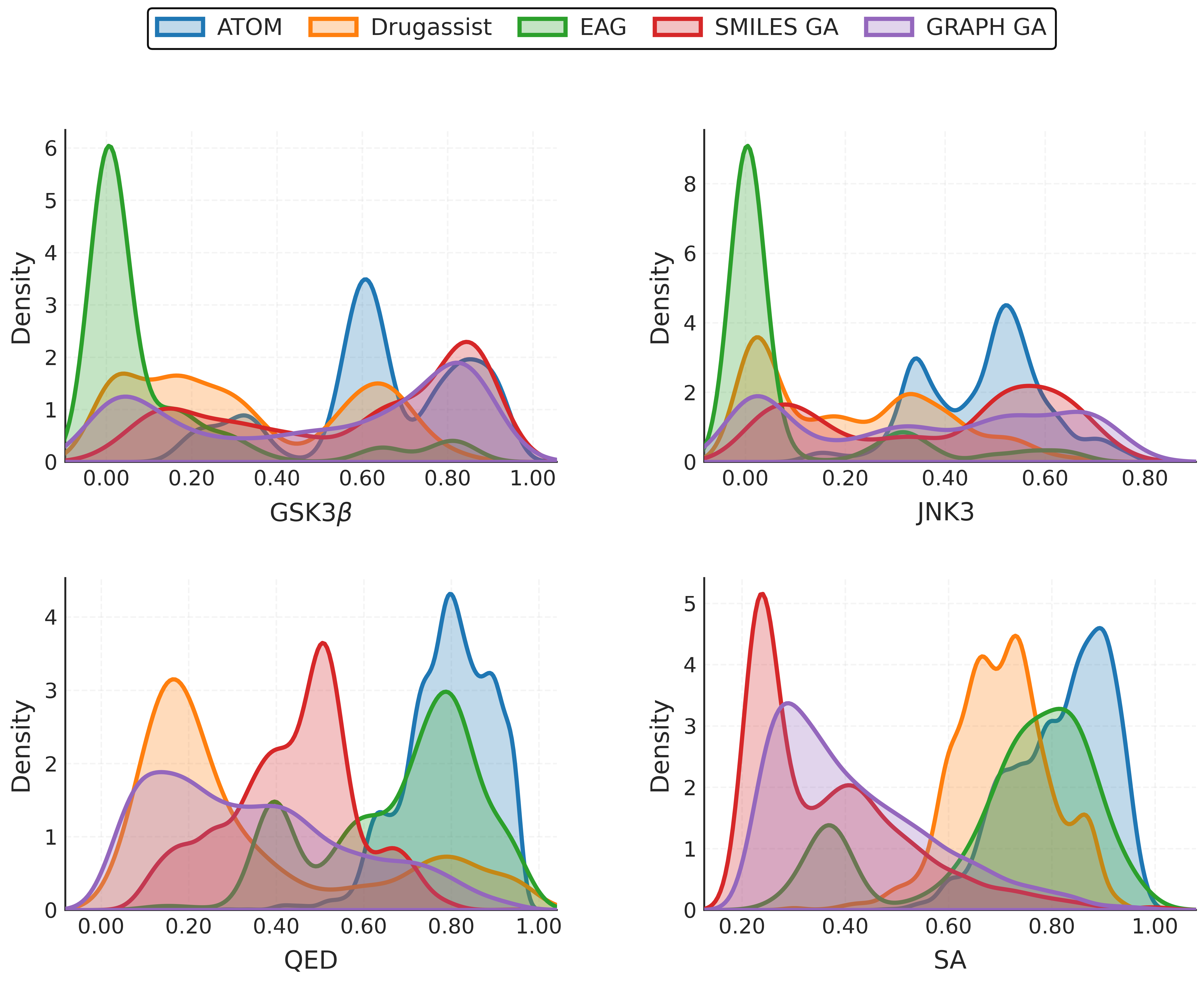}
  \caption{Normalized distributions of the generated molecules, QED and SA GSK3$\beta$ and JNK3.}
  \label{fig:Normalized distributions}
\end{figure}
\textbf{Multi-Objective Results and Pareto Front Analysis.}
To evaluate the optimization efficiency and trade-off management of our proposed method, we analyze the property distributions and Pareto frontiers of the generated molecules. As illustrated in Figure~\ref{fig:Normalized distributions}, ATOM consistently shifts the probability density of key molecular properties—including GSK3$\beta$, JNK3, QED, and SA—towards the optimal high-score regions. While baseline methods such as SMILES GA and GRAPH GA often exhibit inconsistent performance or suffer from a high density of low-quality candidates, ATOM demonstrates a robust ability to identify high-quality molecules across all metrics.This distributional superiority further translates into a more effective exploration of the objective space. 
As shown in the Pareto front analysis in Figure~\ref{fig2:example}, ATOM yields non-dominated solution sets that demonstrate competitive convergence and superior trade-off balance, particularly in drug-likeness-oriented tasks. In the challenging multi-objective scenarios of GSK3$\beta$ + QED and JNK3 + QED, ATOM significantly outperforms baselines by maintaining high biological activity scores ($>0.8$) even as the QED values approach the high-quality region ($\sim 0.9$). While competing other LLM-based methods such as Drugassist and EAG experience a sharp performance decay in target affinity when optimizing for molecular properties, ATOM effectively maintains a broader coverage near the "knee region" of the frontier. These results validate the effectiveness of ATOM’s dynamic coordination mechanism, which adaptively manages the intricate interdependencies between high-dimensional objectives to preserve a more optimal and robust solution distribution.

\begin{figure*}[htbp]
  \centering
  \includegraphics[width=0.9\textwidth]{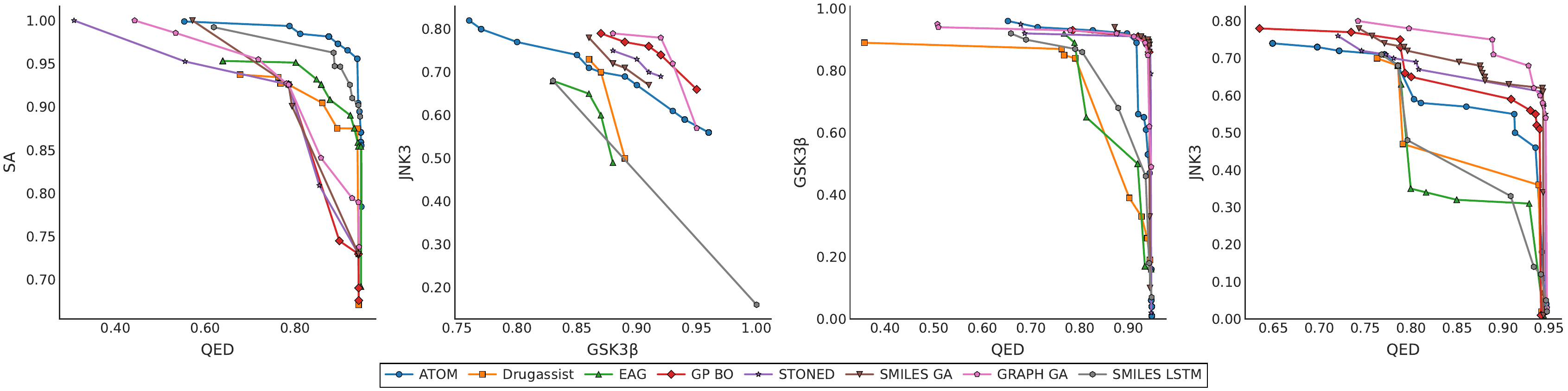}
  \caption{Non-dominated solutions of various methods on GSK3$\beta$+JNK3, QED+SA, GSK3$\beta$+QED and JNK3+QED objectives.}
  \label{fig2:example}
\end{figure*}

\section{Case Studies}

\subsection{Visualization of potential dual-inhibitors for JNK3-GSK3$\beta$}
\begin{figure}[htbp]
  \centering
  \includegraphics[width=1\linewidth]{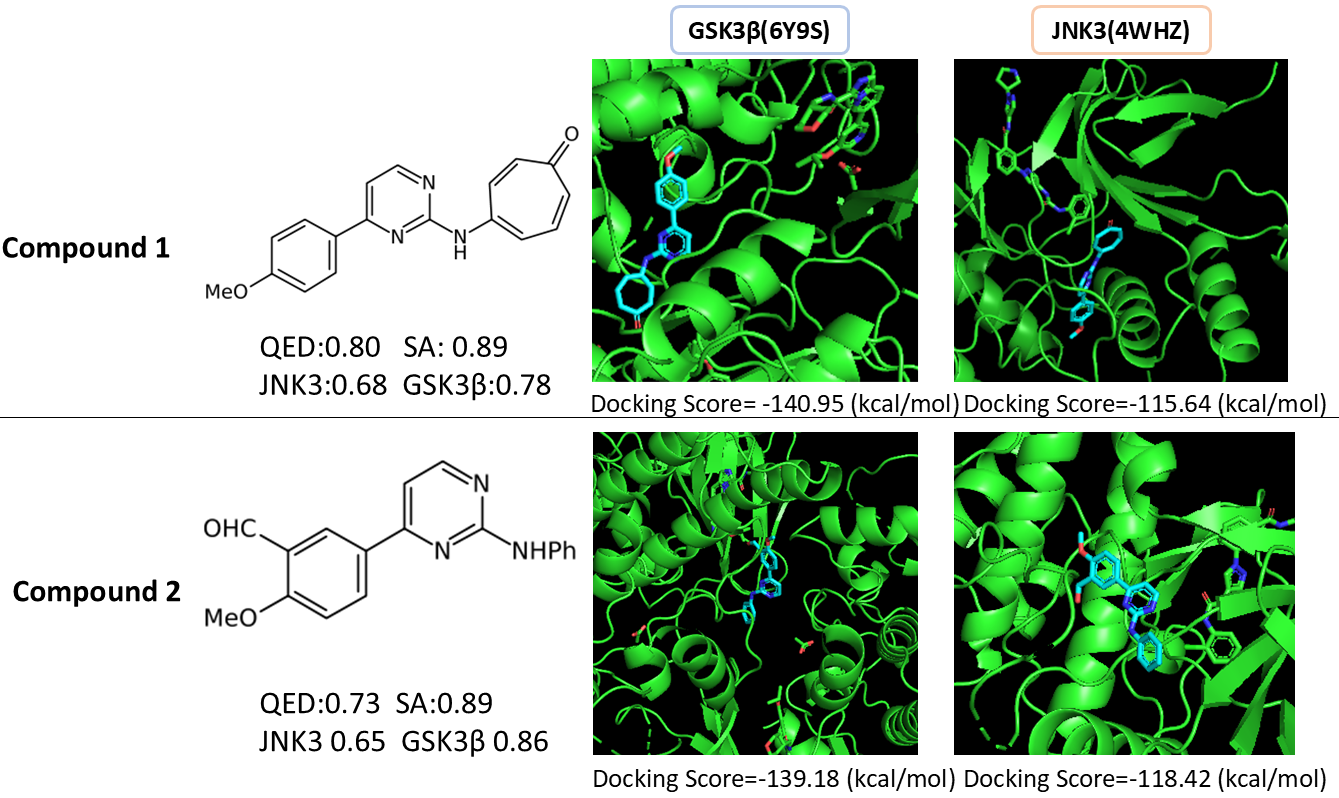}
  \caption{Examples of molecules generated by ATOM on the JNK3-GSK3$\beta$ target pair.}
  \label{fig:case_study_new_2}
\end{figure}
Figure~\ref{fig:case_study_new_2} presents two representative dual-target inhibitors (Compound 1 and Compound 2) generated by ATOM to jointly optimize binding to GSK3$\beta$ (PDB: 6Y9S) and JNK3 (PDB: 4WHZ). 
Docking analysis indicates that both molecules stably occupy the ATP-binding pockets of their respective targets and reproduce canonical kinase interaction motifs reported in prior structural studies, demonstrating ATOM’s ability to learn and transfer conserved binding patterns across targets.

Compound 1 achieved strong predicted affinities for both kinases ($-140.95\ \mathrm{kcal\cdot mol^{-1}}$ for GSK3$\beta$; $-115.64\ \mathrm{kcal\cdot mol^{-1}}$ for JNK3). 
Its binding mode is characterized by hinge hydrogen bonds to VAL135 (GSK3$\beta$) and MET149 (JNK3), complemented by extensive hydrophobic interactions with VAL70 and LEU132 in GSK3$\beta$ and ILE70 and VAL78 in JNK3. 
In addition, a halogen bond with LEU206 in JNK3 further stabilizes the complex, highlighting ATOM’s capacity to exploit target-specific auxiliary interactions.

Compound 2 shows comparable dual-target potency ($-139.18\ \mathrm{kcal\cdot mol^{-1}}$ for GSK3$\beta$; $-118.42\ \mathrm{kcal\cdot mol^{-1}}$ for JNK3). 
Binding is driven by conserved hinge interactions with VAL135 (GSK3$\beta$) and dual hydrogen bonding to MET149 (JNK3), together with hydrophobic packing involving ILE62 (GSK3$\beta$) and VAL78 (JNK3). 
Notably, Compound 2 uniquely forms metal-mediated coordination with a cerium ion in the GSK3$\beta$ active site, illustrating ATOM’s flexibility in accommodating noncanonical yet energetically favorable interaction patterns.

Beyond binding affinity, both compounds exhibit balanced drug-like properties and low synthetic complexity, indicating that ATOM jointly optimizes binding performance and chemical feasibility. Collectively, these results demonstrate that ATOM can generate dual-target kinase inhibitors that preserve key structural interaction motifs while satisfying fundamental drug discovery constraints.
\subsection{Interpretable OptimizationPath}
\begin{figure}[htbp]
  \centering
  \includegraphics[width=0.8\linewidth]{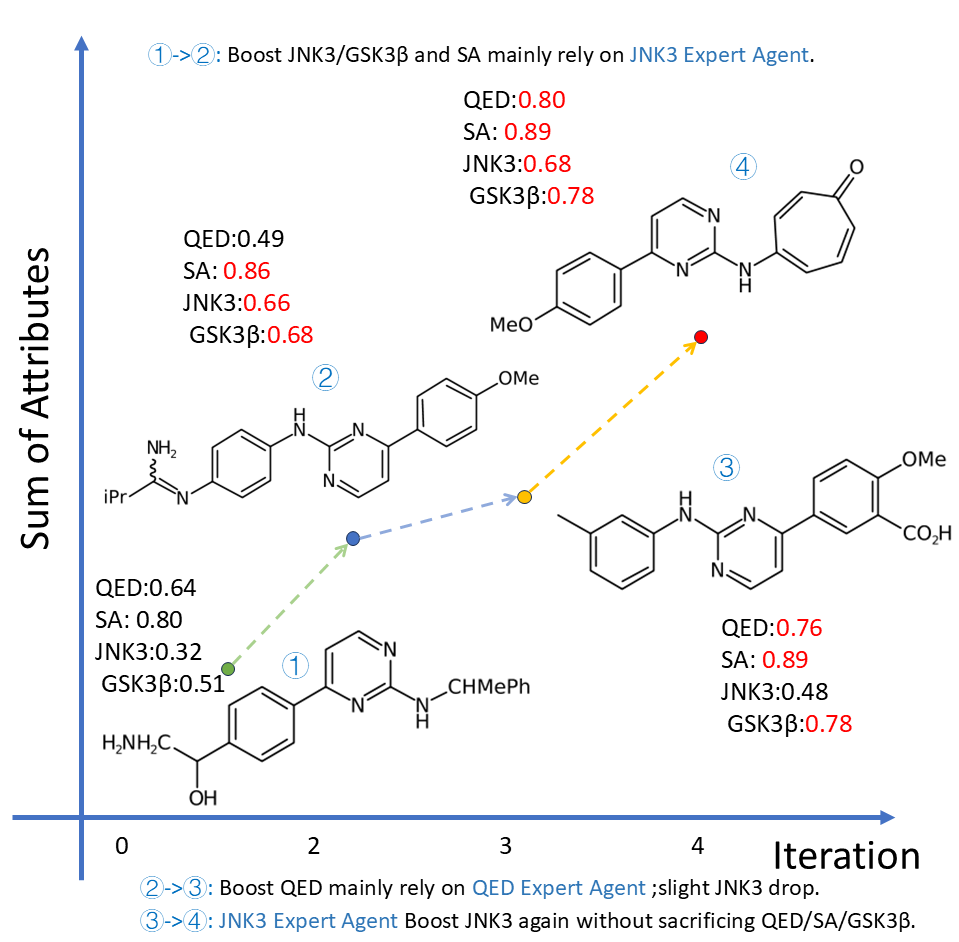}
  \caption{Molecular Optimization Trajectory with ATOM.}
  \label{fig:case_study2}
\end{figure}

Figure~\ref{fig:case_study2} shows a multi-step optimization trajectory starting from a benzene-centered scaffold bearing a heteroaryl amine and a polar amino-alcohol side chain (initial SMILES: NCC(O)c1ccc(-c2ccnc(NC(C)c3ccccc3)n2)cc1). ATOM applies targeted, structure-aware edits guided by atomic attribution maps: early modifications reorganize the polar side chain into a more rigid, amide-like motif and tune peripheral substituents to increase hydrogen-bond directionality, improving predicted activity against both JNK3 and GSK3$\beta$ while only modestly increasing structural complexity; subsequent edits simplify highly basic functionality and introduce a carboxylate to rebalance lipophilicity and reduce overall complexity; finally, a JNK3-specialized agent electronically withdraws the terminal phenyl ring to enhance complementarity in the JNK3 pocket, restoring potency without degrading QED or synthetic accessibility. This trajectory demonstrates that ATOM attains balanced, explainable multi-objective optimization via fine-grained, chemically coherent substituent and functional-group modifications rather than coarse scaffold replacement.

\section*{Conclusion}

We introduced ATOM, a tree-structured multi-agent framework for multi-objective molecular optimization that formulates molecular design as pathwise coordination over alternative evolution trajectories. By assigning specialized agents to atomic operations along different branches and integrating a global memory for cross-path information sharing, ATOM effectively captures conflicting trade-offs and long-horizon dependencies in chemical spaces. Extensive experiments on challenging benchmarks demonstrate consistent improvements in Pareto coverage and hypervolume over strong baselines, highlighting the promise of tree-structured multi-agent coordination for complex molecular optimization tasks.

\section*{Impact Statement}
This paper presents work whose goal is to advance the field of machine learning. There are many potential societal consequences of our work, none of which we feel must be specifically highlighted here.

\nocite{langley00}

\bibliography{example_paper}
\bibliographystyle{icml2026}

\newpage
\appendix
\onecolumn
\section{Expert Prompt Design}

This section provides a detailed description of the prompt design templates for each expert agent in the ATOM framework, as well as the auxiliary tool components employed during optimization. When constructing prompts, we closely follow the example formats provided in the original open-source implementations to ensure that the large language model (LLM) can reliably interpret task instructions and execute the specified operations correctly. To ensure stable system behavior and facilitate downstream parsing, the output format of each expert agent is strictly standardized.

Figures~7, 8, and~9 present the prompt templates used for the GSK3$\beta$ expert, the JNK3 expert, and the synthetic accessibility (SA) expert, respectively. For the SA property, we apply a normalization procedure to rescale its value range to $[0,1]$ and reformulate the original minimization objective as a maximization problem. Specifically, an SA value closer to~1 indicates higher synthetic accessibility, whereas values closer to~0 correspond to increased synthetic difficulty.

The ATOM framework integrates two primary tool components: \textbf{RDKit} and \textbf{Oracle}. RDKit is a widely used open-source cheminformatics library that supports molecular representation, manipulation, and analysis across multiple formats, including SMILES, SMARTS, and SDF. It serves as the core backend for molecular construction and property computation during the optimization process. The Oracle library provides a unified interface for molecular property evaluation and similarity assessment, enabling efficient screening of candidate molecules based on predicted properties and structural similarity. Together, these tools support reliable and scalable molecular evaluation within the ATOM framework.


\begin{figure}[htbp]
\centering
\begin{tcolorbox}[
  colback=green!5!white,
  colframe=green!60!black,
  title=\textbf{GSK3$\beta$ Expert Prompt},
  fonttitle=\bfseries,
  boxrule=0.8pt,
  arc=6pt,
  left=6pt,
  right=6pt,
  top=6pt,
  bottom=6pt
]

\textbf{Role Description:} \\
You are an expert computational chemist and molecular optimization specialist.
You have extensive knowledge of medicinal chemistryotation, and structure–activity relationships (SAR).
Your task is to design novel molecules that show improved GSK3beta inhibitory activity while maintaining good drug-likeness and chemical validity.
Your goal is to optimize these molecules to generate 50 new compounds with improved GSK3beta inhibition activity.

Please follow these guidelines carefully:

Input: A list of starting SMILES strings (example below).

Objective: Increase the predicted or expected GSK3beta inhibitory activity.

Output: Generate exactly 50 valid SMILES strings representing new molecules.

\vspace{0.5em}
\textbf{Constraints:} \\
The generated molecules should retain the core scaffold or key pharmacophores of the starting compounds (do not generate entirely unrelated structures).
Ensure synthetic accessibility (reasonable SA scores) and good drug-likeness (high QED or Lipinski compliance).
Each SMILES must be syntactically valid and parseable by RDKit.

\vspace{0.5em}
\textbf{Optimization strategies} \\
\{Adjusted according to reference molecular dynamics\}

\vspace{0.5em}
\textbf{reference molecular:} \\
\{Select from parent node, sibling nodes, and dynamic knowledge base\}

\vspace{0.5em}
\textbf{Output Requirements:} \\
Output the 50 optimized SMILES only, one per line, numbered from 1 to 50.
Do not include explanations, commentary, or non-SMILES text.

\end{tcolorbox}

\caption{Prompt design template for the GSK3$\beta$ expert agent.}
\end{figure}

\begin{figure}[htbp]
\centering
\begin{tcolorbox}[
  colback=green!5!white,
  colframe=green!60!black,
  title=\textbf{JNK3 Expert Prompt},
  fonttitle=\bfseries,
  boxrule=0.8pt,
  arc=6pt,
  left=6pt,
  right=6pt,
  top=6pt,
  bottom=6pt
]

\textbf{Role Description:} \\
You are an expert computational chemist and molecular optimization specialist.
You have extensive knowledge of medicinal chemistryotation, and structure–activity relationships (SAR).
Your task is to design novel molecules that show improved JNK3 inhibitory activity while maintaining good drug-likeness and chemical validity.
Your goal is to optimize these molecules to generate 50 new compounds with improved JNK3 inhibition activity.

Please follow these guidelines carefully:

Input: A list of starting SMILES strings (example below).

Objective: Increase the predicted or expected JNK3 inhibitory activity.

Output: Generate exactly 50 valid SMILES strings representing new molecules.

\vspace{0.5em}
\textbf{Constraints:} \\
The generated molecules should retain the core scaffold or key pharmacophores of the starting compounds (do not generate entirely unrelated structures).
Ensure synthetic accessibility (reasonable SA scores) and good drug-likeness (high QED or Lipinski compliance).
Each SMILES must be syntactically valid and parseable by RDKit.

\vspace{0.5em}
\textbf{Optimization strategies} \\
\{Adjusted according to reference molecular dynamics\}

\vspace{0.5em}
\textbf{reference molecular:} \\
\{Select from parent node, sibling nodes, and dynamic knowledge base\}

\vspace{0.5em}
\textbf{Output Requirements:} \\
Output the 50 optimized SMILES only, one per line, numbered from 1 to 50.
Do not include explanations, commentary, or non-SMILES text.

\end{tcolorbox}
\end{figure}

\begin{figure}[htbp]
\centering
\begin{tcolorbox}[
  colback=green!5!white,
  colframe=green!60!black,
  title=\textbf{SA Expert Prompt},
  fonttitle=\bfseries,
  boxrule=0.8pt,
  arc=6pt,
  left=6pt,
  right=6pt,
  top=6pt,
  bottom=6pt
]

\textbf{Role Description:} \\
You are an expert computational chemist and molecular optimization specialist.
You have extensive knowledge of medicinal chemistry, SMILES notation, synthesis planning,and structure–activity relationships (SAR).
Your task is to design novel molecules that improve synthetic accessibility (lower SA score) while maintaining stable JNK3 and GSK3beta activities and preserving the drug-like properties.

Please follow these guidelines carefully:

Input: A list of starting SMILES strings (example below).

Objective: Decrease the predicted SA score (make the molecule easier to synthesize)

Output: Generate exactly 50 valid SMILES strings representing new molecules.

\vspace{0.5em}
\textbf{Constraints:} \\
Exactly 50 output SMILES
Each SMILES must be valid and RDKit-parseable
Modifications should be mild and chemically meaningful:simplify overly complex ring linkages,replace exotic fragments with classical synthetic handles
,reduce steric congestion or macrocyclic motifs
,use robust bioisosteric substitutions
,avoid exotic heterocycles or unusual protecting-group-like fragments.

\vspace{0.5em}
\textbf{Optimization strategies} \\
\{Adjusted according to reference molecular dynamics\}

\vspace{0.5em}
\textbf{reference molecular:} \\
\{Select from parent node, sibling nodes, and dynamic knowledge base\}

\vspace{0.5em}
\textbf{Output Requirements:} \\
Output the 50 optimized SMILES only, one per line, numbered from 1 to 50.
Do not include explanations, commentary, or non-SMILES text.

\end{tcolorbox}

\caption{Prompt design template for the SA expert agent.}
\end{figure}

\section{Introduction To Related Properties}
We consider the following molecular objectives used throughout this work: inhibition potency against GSK3\(\beta\) and JNK3, and two widely used drug-likeness / synthesizability metrics (QED and SA).

\paragraph{GSK3\(\beta\).}
Glycogen synthase kinase 3 beta (\(\mathrm{GSK3}\beta\)) is a serine/threonine protein kinase involved in glycogen metabolism, cell proliferation, differentiation, and apoptosis. It has attracted substantial interest in neurodegenerative disease research because it regulates tau phosphorylation, amyloid precursor protein processing, and neuronal survival. Consequently, GSK3\(\beta\) inhibition is considered an important molecular objective in drug discovery targeting neurodegeneration.

\paragraph{JNK3.}
c-Jun N-terminal kinase 3 (JNK3) is a member of the MAPK family primarily expressed in the central nervous system. JNK3 mediates cellular stress responses including apoptosis and inflammation; inhibiting JNK3 is therefore a relevant objective for mitigating neuronal cell death and inflammatory processes in neurodegenerative disease models.

\paragraph{QED.}
Quantitative Estimate of Drug-likeness (QED) aggregates several physicochemical properties (molecular weight, lipophilicity/logP, topological polar surface area, counts of H-bond donors/acceptors, aromatic ring count, and rotatable bonds) into a single score on the interval \([0,1]\). Higher QED values indicate molecules that are more likely to exhibit favorable drug-like properties.

\paragraph{SA and normalized SA.}
Synthetic Accessibility (SA) is a heuristic score used to estimate the ease of laboratory synthesis; SA typically ranges from 1 (easy) to 10 (difficult). To treat synthetic feasibility as a maximization objective, we follow prior work and convert SA to a normalized score in \([0,1]\) via
\begin{equation}
\mathrm{Normalized\_SA} \;=\; 1 - \frac{\mathrm{SA} - 1}{9},
\label{eq:normalized_sa}
\end{equation}
so that larger values correspond to higher synthetic accessibility.

\section{Details of correlation calculation between different objectives}
To characterize the relationships among the molecular objectives considered in this work, we conducted a correlation analysis on a random subset of molecules sampled from the ZINC20 database. Specifically, we evaluated the pairwise dependencies between drug-likeness (QED), synthetic accessibility (SA), and predicted binding affinities to JNK3 and GSK3$\beta$.

For each molecule, QED and SA scores were computed using RDKit, while JNK3 and GSK3$\beta$ scores were obtained from the corresponding oracle models. We then computed pairwise Spearman rank correlation coefficients, which capture monotonic relationships without assuming linearity and are robust to non-Gaussian score distributions commonly observed in molecular property spaces.

Figure~\ref{fig:example} presents the Pearson correlation coefficients between optimization objectives. We observe that QED and SA are nearly decoupled ($r = 0.051$), while the two kinase objectives (GSK3$\beta$ and JNK3) show a moderate positive correlation ($r = 0.351$), likely reflecting shared binding motifs. The consistently weak correlations between biological activity and chemical feasibility metrics ($|r| \leq 0.220$) suggest that improvements in binding affinity do not inherently translate to superior drug-likeness, confirming the necessity of a multi-objective optimization approach.

These results highlight the intrinsic multi-objective nature of the optimization problem and motivate the use of explicit multi-objective optimization strategies in this work.

\section{Performance on Top-ranked Candidate Molecules}
We further evaluate method effectiveness by reporting top-10 and top-50 average scores for each objective, together with overall average score and average rank (Tables~2 and~3).
For the biological objective pair (GSK3$\beta$ + JNK3), ATOM achieves the best overall ranking (average rank = 1.0), with consistently strong top-10 and top-50 scores on both targets (GSK3$\beta$: 0.953 / 0943; JNK3: 0.792 / 0.754). In contrast, while STONED performs well on GSK3$\beta$ (top-50: 0.897), its performance on JNK3 is substantially lower (average: 0.342), indicating difficulty in jointly optimizing correlated bioactivity objectives. These results highlight ATOM's advantage in coordinating expert agents under interdependent biological constraints.
For the non-biological task (QED + SA), multiple methods achieve competitive results. ATOM attains the highest average QED score (0.794) and SA score (0.819). SMILES GA slightly outperforms ATOM on top-10 SA (0.997) but lags behind on QED. Overall, these results suggest that while some baselines excel at individual non-biological objectives, ATOM maintains more balanced performance across objectives, particularly in challenging biological settings.

\begin{table*}[htbp]
\centering
\label{tab:method-comparison-jnk-gsk}
\small
\begin{tabular}{lccccccc}
\toprule
Method & Top10 GSK3$\beta$ & Top10 JNK3 & Top50 GSK3$\beta$ & Top50 JNK3 & Avg GSK3$\beta$ & Avg JNK3  & Avg Rank\\
\midrule
SMILES\ LSTM & 0.840 & 0.608 & 0.678 & 0.428 & 0.090 & 0.040 & 9\\
SMILES\ GA        & 0.906 & 0.763 & 0.896 & 0.729 & 0.572 & 0.388 & 3\\
GRAPH\ GA        & 0.927 & 0.772 & 0.902 & 0.753 & 0.501 & 0.354 & 4\\
STONED            & 0.910 & 0.738 & 0.897 & 0.726 & 0.342 & 0.207 & 5\\
GP BO             & 0.932 & 0.787 & 0.914 & 0.776 & 0.764 & 0.541 & 2\\
Drugassist        & 0.860 & 0.688 & 0.810 & 0.625 & 0.330 & 0.211 & 6\\
GPT4o-mini   & 0.841 & 0.704 & 0.834 & 0.685 & 0.151 & 0.085 & 7\\
EAG               & 0.871 & 0.680 & 0.847 & 0.658 & 0.120 & 0.080 & 8\\
ATOM              & 0.952 & 0.792 & 0.943 & 0.754 & 0.765 & 0.543 & 1 \\
\bottomrule
\end{tabular}
\caption{Comparison of GSK3$\beta$ and JNK3 performance across different methods.}
\end{table*}

\begin{table*}[htbp]
\centering
\small
\begin{tabular}{lccccccc}
\toprule
Method & Top10 Avg QED & Top10 Avg SA & Top50 Avg QED & Top50 Avg SA & Avg QED & Avg SA & Avg Rank\\
\midrule
SMILES\ LSTM & 0.940 & 0.953 & 0.925 & 0.926 & 0.572 & 0.746 & 4\\
SMILES\ GA   & 0.814 & 0.997 & 0.762 & 0.880 & 0.437 & 0.369 & 5\\
GRAPH\ GA   & 0.933 & 0.958 & 0.890 & 0.879 & 0.346 & 0.418 & 8\\
STONED       & 0.855 & 0.910 & 0.794 & 0.786 & 0.379 & 0.401 & 9\\
GPBO         & 0.924 & 0.806 & 0.883 & 0.724 & 0.551 & 0.524 & 6\\
Drugassist   & 0.941 & 0.930 & 0.941 & 0.901 & 0.339 & 0.703 & 7\\
GPT4o-mini   & 0.944 & 0.937 & 0.942 & 0.931 & 0.675 & 0.744 & 3\\
EAG          & 0.943 & 0.938 & 0.938 & 0.938 & 0.683 & 0.711 & 2\\
ATOM         & 0.947 & 0.985 & 0.946 & 0.977 & 0.794 & 0.819 & 1\\
\bottomrule
\end{tabular}
\caption{Comparison of QED and SA performance across different methods.}
\label{tab:qed_sa}
\end{table*}

\section{Diversity and Novelty Analysis}

In Table~4, we present the diversity (Div) and novelty (Nov) scores of molecular libraries generated by different methods under six multi-objective optimization settings. Among them, the \textbf{ATOM} method demonstrates consistently strong and balanced performance across nearly all experimental configurations.
Notably, ATOM achieves exceptionally high novelty scores, reaching 0.9994 or above in most settings and approaching or attaining values close to 0.9999 in several cases. This substantially outperforms many conventional generative models (e.g., SMILES LSTM, GB PO, Druggassist) and several reinforcement-learning or evolutionary algorithm baselines, indicating that ATOM is highly effective at exploring previously under-sampled regions of chemical space and generating structurally novel molecules.
At the same time, ATOM maintains competitive diversity scores, typically falling in the 0.75--0.85 range across most settings. In certain combinations (e.g., QED+SA, JNK3+QED+SA), its diversity is comparable to or even superior to some strong baselines, showing that it can preserve reasonable molecular diversity while satisfying drug-likeness and synthesizability constraints.
Overall, ATOM achieves a favorable trade-off between diversity and novelty in multi-objective molecule generation. Its ability to deliver extremely high novelty while still maintaining acceptable diversity makes it particularly promising for drug discovery scenarios that require both exploration of novel chemical matter and adherence to desired property distributions.

\begin{table*}[htbp]
\centering
\scriptsize
\setlength{\tabcolsep}{3.5pt}  

\label{tab:multi-obj-div-nov-compact}
\begin{tabular}{l *{12}{c}}
\toprule
Method & \multicolumn{2}{c}{QED+SA} & \multicolumn{2}{c}{GSK3$\beta$+JNK3} & \multicolumn{2}{c}{QED+JNK3+GSK3$\beta$} & \multicolumn{2}{c}{GSK3$\beta$+QED+SA} & \multicolumn{2}{c}{JNK3+QED+SA} & \multicolumn{2}{c}{QED+JNK3+GSK3$\beta$+SA} \\
\cmidrule(lr){2-3} \cmidrule(lr){4-5} \cmidrule(lr){6-7} \cmidrule(lr){8-9} \cmidrule(lr){10-11} \cmidrule(lr){12-13}
       & Div    & Nov    & Div    & Nov    & Div    & Nov    & Div    & Nov    & Div    & Nov    & Div    & Nov \\
\midrule
SMILES\ LSTM & 0.8833 & 0.9996 & 0.8799 & 0.9994 & 0.8780 & 0.9994 & 0.8780 & 0.9994 & 0.8800 & 0.9994 & 0.8800 & 0.9994 \\
SMILES\ GA   & 0.8555 & 0.9994 & 0.7465 & 0.9994 & 0.7444 & 0.9996 & 0.8253 & 0.9997 & 0.8689 & 0.9997 & 0.7776 & 0.9997 \\
GRAPH\ GA    & 0.8690 & 0.9997 & 0.7717 & 0.9995 & 0.8162 & 0.9997 & 0.8518 & 0.9997 & 0.8899 & 0.9997 & 0.8485 & 0.9997 \\
STONED       & 0.7700 & 0.9996 & 0.8313 & 0.9997 & 0.8389 & 0.9997 & 0.7828 & 0.9997 & 0.9063 & 0.9991 & 0.7878 & 0.9996 \\
GB PO        & 0.5944 & 0.9967 & 0.6721 & 0.9967 & 0.6560 & 0.9967 & 0.6253 & 0.9967 & 0.6000 & 0.9967 & 0.5475 & 0.9967 \\
GPT4o-mini    & 0.7914 & 0.9667 & 0.8067 & 0.9778 & 0.7911 & 0.9720 & 0.6890 & 0.9667 & 0.8101 & 0.9745 & 0.8040 & 0.9750 \\
Drugassist   & 0.7544 & 0.9934 & 0.7507 & 0.9934 & 0.7553 & 0.9934 & 0.7501 & 0.9934 & 0.7511 & 0.9934 & 0.7655 & 0.9934 \\
EAG          & 0.8445 & 0.9766 & 0.8045 & 0.9900 & 0.8244 & 0.9899 & 0.8401 & 0.9912 & 0.8341 & 0.9911 & 0.8130 & 0.9912 \\
ATOM         & 0.8394 & 0.9997 & 0.7020 & 0.9999 & 0.8182 & 0.9991 & 0.8278 & 0.9998 & 0.8482 & 0.9998 & 0.7521 & 0.9994 \\

\bottomrule
\end{tabular}
\caption{Diversity (Div) and Novelty (Nov) of generated molecules under different multi-objective settings. All values are rounded to 4 decimal places.}

\end{table*}

\section{Assumption of Theoretical Analysis}

\textbf{Assumption 1} (Orthogonal Descent Capability). For any state $s$ that is a local optimum for objective $i$ (where $\nabla f_i(s) \approx 0$) but not Pareto optimal, there exists at least one agent $Ag_j$ ($j \neq i$) such that the expected improvement in the joint value function is lower-bounded\cite{van2013scalarized,alegre2023sample}:
\begin{equation}
\mathbb{E}[V(T(s, Ag_j)) - V(s)] \ge \delta > 0.
\label{eq3:transition} 
\end{equation}
This implies that the agents provide "orthogonal" gradient directions to escape single-objective local optima\cite{suzuki2024mothra}.

\textbf{Assumption 2} (Optimistic Pruning Condition). The exploration bonus $c_d$ is calibrated such that with high probability $1-\beta$, the true value $V(s)$ is bounded by the optimistic estimate\cite{munos2014bandits,kocsis2006bandit}:
\begin{equation}
|V(s) - U(s)| \le c_d.
\label{eq4:transition} 
\end{equation}
Furthermore, the noise $\sigma_d$ (and thus $c_d$) decays as a function of depth $d$ (e.g., $\sigma_d \propto d^{-\gamma}$ or $e^{-\alpha d}$), reflecting that evaluations become more certain closer to leaf nodes (refined molecules).

\textbf{Assumption 3} (Knowledge Base Guidance). The Dynamic Knowledge Base $\mathcal{K}$ reduces the effective search horizon\cite{zhang2025rag2mol,wang2022retrieval}. For a target property profile requiring $L$ edit steps from $s_0$, the conditional generation probability using KB prompts satisfies:
\begin{equation}
\mathbb{P}(\text{target in } L_{KB} \text{ steps} \mid \mathcal{K}) \ge \mathbb{P}(\text{target in } L \text{ steps} \mid \emptyset),
\label{eq6:transition} 
\end{equation}
where $L_{KB} < L$. This effectively contracts the required search depth for complex objectives.

\section{Baseline Details}

\label{app:baselines}

In this work, we compare our method against several established baselines on molecular optimization tasks. 
The implementations used are listed below, prioritizing official code repositories or standardized benchmark suites wherever available.

\begin{itemize}
    \item \textbf{SMILES LSTM} \\
    A sequence-based LSTM model operating on SMILES strings, typically combined with hill-climbing or reinforcement learning strategies for property optimization. \\
    We use the official implementation from the ChemLactica Test Suite: \\
    \url{https://github.com/YerevaNN/ChemLacticaTestSuite/tree/master/mol_opt} 
    (specifically \texttt{smiles\_lstm\_hc}).

    \item \textbf{SMILES GA} \\
    A genetic algorithm operating directly on SMILES strings, performing crossover and mutation in string space. \\
    We use the official implementation from the ChemLactica Test Suite: \\
    \url{https://github.com/YerevaNN/ChemLacticaTestSuite/tree/master/mol_opt} 
    (specifically \texttt{smiles\_ga}).

    \item \textbf{Graph GA} \\
    A genetic algorithm that operates on molecular graph representations, enabling more chemically meaningful edit operations. \\
    We use the official implementation from the ChemLactica Test Suite: \\
    \url{https://github.com/YerevaNN/ChemLacticaTestSuite/tree/master/mol_opt} 
    (specifically \texttt{graph\_ga}).

    \item \textbf{STONED} \\
    A fragment-based genetic algorithm using SELFIES representations for efficient local exploration of chemical space. \\
    We use the official implementation from the ChemLactica Test Suite: \\
    \url{https://github.com/YerevaNN/ChemLacticaTestSuite/tree/master/mol_opt} 
    (specifically \texttt{stoned}).

    \item \textbf{GB PO} \\
    A graph-based or Bayesian optimization method for property-guided molecular improvement (as implemented in the benchmark suite, closely aligned with GPBO-style approaches). \\
    We use the official implementation from the ChemLactica Test Suite: \\
    \url{https://github.com/YerevaNN/ChemLacticaTestSuite/tree/master/mol_opt}.

    \item \textbf{DrugAssist} \\
    A large language model-based approach for molecule optimization guided by natural language prompts. \\
    We use the official code repository: \\
    \url{https://github.com/blazerye/DrugAssist}.

    \item \textbf{EAG} \\
    A sequential multi-agent collaboration method for complex reasoning (details based on the original description).  \\
    Since no official open-source implementation is available, we re-implemented the core components based on the method's published algorithmic description.
\end{itemize}

\end{document}